\documentclass[10pt,twocolumn,letterpaper]{article}

\usepackage{iccv}
\usepackage{times}
\usepackage{epsfig}
\usepackage{graphicx}
\usepackage{amsmath}
\usepackage{amssymb}
\usepackage{tabularx} 
\usepackage[lofdepth,lotdepth]{subfig}
\usepackage{enumitem}
\usepackage{booktabs}
\usepackage{arydshln}
\usepackage{subfig}
\usepackage{color}
\usepackage[switch]{lineno}
\usepackage{float}
\usepackage[numbers,sort&compress]{natbib} 
\usepackage[symbol]{footmisc}

\definecolor{arthurcolor}{RGB}{0, 80, 160}
\definecolor{bingcolor}{RGB}{0, 160, 80}
\definecolor{yanyicolor}{RGB}{255, 97, 0}
\definecolor{biagiocolor}{RGB}{160, 80, 0}

\usepackage[pagebackref=true,breaklinks=true,letterpaper=true,colorlinks,bookmarks=false]{hyperref}

\iccvfinalcopy 

\def\iccvPaperID{} 

\ificcvfinal\fi

\begin{document}

\title{VidTr: Video Transformer Without Convolutions}

\author{Yanyi Zhang \textsuperscript{\rm 1,2} \thanks{Equally Contributed.}, Xinyu Li \textsuperscript{\rm 1}\footnotemark[1], Chunhui Liu \textsuperscript{\rm 1}, Bing Shuai \textsuperscript{\rm 1}, Yi Zhu \textsuperscript{\rm 1}, \\Biagio Brattoli \textsuperscript{\rm 1}, Hao Chen \textsuperscript{\rm 1}, Ivan Marsic \textsuperscript{\rm 2} and Joseph Tighe \textsuperscript{\rm 1}\\
\textsuperscript{\rm 1} Amazon Web Service; \textsuperscript{\rm 2} Rutgers University\\
{\tt\small \{xxnl,chunhliu,bshuai,yzaws,biagib,hxen,tighej\}@amazon.com; \{yz593,marsic\}@rutgers.edu}
}
 
\maketitle
\ificcvfinal\thispagestyle{empty}\fi

\begin{abstract}
We introduce Video Transformer (VidTr) with separable-attention for video classification. Comparing with commonly used 3D networks, VidTr is able to aggregate spatio-temporal information via stacked attentions and provide better performance with higher efficiency. 
We first introduce the vanilla video transformer and show that transformer module is able to perform spatio-temporal modeling from raw pixels, but with heavy memory usage.
We then present VidTr which reduces the memory cost by 3.3$\times$ while keeping the same performance. 
To further optimize the model, we propose the standard deviation based topK pooling for attention ($pool_{topK\_std}$), which reduces the computation by dropping non-informative features along temporal dimension.
VidTr achieves state-of-the-art performance on five commonly used datasets with lower computational requirement, showing both the efficiency and effectiveness of our design.
Finally, error analysis and visualization show that VidTr is especially good at predicting actions that require long-term temporal reasoning.
\end{abstract}

\section{Introduction}
We introduce Video Transformer (VidTr) with separable-attention, one of the first transformer-based video action classification architecture that performs global spatio-temporal feature aggregation.
Convolution-based architectures have dominated the video classification literature in recent years \cite{wang2018non,feichtenhofer2018slowfast,li2020directional}, and although successful, the convolution-based approaches have two drawbacks: 1. they have limited receptive field on each layer and 2. information is slowly aggregated through stacked convolution layers, which is inefficient and might be ineffective \cite{wang2018non,li2020nuta}. Attention is a potential candidate to overcome these limitations as it has a large receptive field which can be leveraged for spatio-temporal modeling. Previous works use attention to modeling long-range spatio-temporal features in videos but still rely on convoluational backbones \cite{wang2018non,li2020nuta}. Inspired by recent successful applications of transformers on NLP \cite{vaswani2017attention,devlin2018bert} and computer vision \cite{dosovitskiy2020image,touvron2020training}, we propose a transformer-based video network that directly applies attentions on raw video pixels for video classification, aiming at higher efficiency and better performance.

We first introduce a vanilla video transformer that directly learns spatio-temporal features from raw-pixel inputs via vision transformer \cite{dosovitskiy2020image}, showing that it is possible to perform pixel-level spatio-temporal modeling. However, as discussed in \cite{wu2019long}, the transformer has $\mathbb{O}(n^2)$ complexity with respect to the sequence length. The vanilla video transformer is memory consuming, as training on a 16-frame clip ($224\times224$) with only batch size of 1 requires more than 16GB GPU memory, which makes it infeasible on most commercial devices.
Inspired by the R(2+1)D convolution that breaks down 3D convolution kernel to a spatial kernel and a temproal kernel \cite{tran2018closer}, we further introduce our separable-attention, which performs spatial and temporal attention separately. This reduces the memory consumption by 3.3$\times$ with no drop in accuracy. We can further reduce the memory and computational requirements of our system by exploiting the fact that a large portion of many videos have redundant information temporally. This notion has been explored in the context of convolutional networks to reduce computation previously \cite{li2020directional}. We build on this intuition and propose a standard deviation based topK pooling operation ($topK\_std$ pooling), which reduces the sequence length and encourages the transformer network to focus on representative frames.  

We evaluated our VidTr on 6 most commonly used datasets, including Kinetics 400/700, Charades, Something-something V2, UCF-101 and HMDB-51. Our model achieved state-of-the-art (SOTA) or comparable performance on five datasets with lower computational requirements and latency compared to previous SOTA approaches.  
Our error analysis and ablation experiments show that the VidTr works significantly better than I3D on activities that requires longer temporal reasoning (e.g. making a cake vs. eating a cake), which aligns well with our intuition. This also inspires us to ensemble the VidTr with the I3D convolutional network as features from global and local modeling methods should be complementary. 
We show that simply combining the VidTr with a I3D50 model (8 frames input) via ensemble  can lead to roughly a 2\% performance improvement on Kinetics 400.
We further illustrate how and why the VidTr works by visualizing the separable-attention using attention rollout \cite{abnar2020quantifying}, and show that the spatial-attention is able to focus on informative patches while temporal attention is able to reduce the duplicated/non-informative temporal instances. 
Our contributions are: 

\begin{enumerate}[itemsep=0pt,parsep=0pt]
\small
\item \textbf{Video transformer: } We propose to efficiently and effectively aggregate spatio-temporal information with stacked attentions as opposed to convolution based approaches. We introduce vanilla video transformer as proof of concept with SOTA comparable performance on video classification.
\item \textbf{VidTr: } We introduce VidTr and its permutations, including the VidTr with SOTA performance and the compact-VidTr with significantly reduced computational costs using the proposed standard deviation based pooling method.
\item \textbf{Results and model weights: } We provide detailed results and analysis on 6 commonly used datasets which can be used as reference for future research. Our pre-trained model can be used for many down-streaming tasks. 
\end{enumerate}

\section{Related Work}
\subsection{Action Classification}
The early research on video based action recognition relies on 2D convolutions \cite{karpathy2014large}. The LSTM \cite{hochreiter1997long} was later proposed to model the image feature based on ConvNet features \cite{yue2015beyond,ullah2017action,li2016action}. However, the combination of ConvNet and LSTM did not lead to significantly better performance. Instead of relying on RNNs, the segment based method TSN  \cite{wang_ECCV2016_TSN} and its permutations \cite{girdhar_CVPR2017_actionVLAD,Li_CVPR2016_VLAD3,zhou_ECCV2018_TRN} were proposed with good performance.

Although 2D network was proved successful, the spatio-temporal modeling was still separated. Using 3D convolution for spatio-temporal modeling was initially proposed in \cite{ji20123d} and further extended to the C3D network \cite{tran2015learning}. However, training 3D convnet from scratch was hard, initializing the 3D convnet weights by inflate from 2D networks was initially proposed in I3D \cite{carreira2017quo} and soon proved applicable with different type of 2D network \cite{hara2018can,chen2018multi,xie2017aggregated}.
The I3D was used as backbone for many following work including two-stream network \cite{wang2018non,feichtenhofer2018slowfast}, the networks with focus on temporal modeling \cite{li2020directional, li2020nuta,yang_CVPR2020_TPN}, and the 3D networks with refined 3D convolution kernels \cite{jiang_ICCV2019_STM,li_CVPR2020_TEA,liu_AAAI2020_TEINet,shao2020temporal}.

The 3D networks are proved effective but often not efficient, the 3D networks with better performance often requires larger kernels or deeper structures. The recent research demonstrates that depth convolution significantly reduce the computation \cite{tran2019video}, but depth convolution also increase the network inference latency. TSM \cite{lin2019tsm} and TAM \cite{fan2019more} proposed a more efficient backbone for temporal modeling, however, such design couldn't achieve SOTA performance on Kinetics dataset. The neural architecture search was proposed for action recognition \cite{feichtenhofer2020x3d,piergiovanni2019tiny} recently with competitive performance, however, the high latency and limited generalizability remain to be improved.

The previous methods heavily rely on convolution to aggregate features spatio-temporally, which is not efficient. A few previous work tried to perform global spatio-temporal modeling \cite{wang2018non,li2020nuta} but still limited by the convolution backbone. The proposed VidTr is fundamentally different from previous works based on convolutions, the VidTr doesn't require heavily stacked convolutions \cite{yang_CVPR2020_TPN} for feature aggregation but efficiently learn feature globally via attention from first layer. Besides, the VidTr don't rely on sliding convolutions and depth convolutions, which runs at less FLOPs and lower latency compared with 3D convolutions \cite{yang_CVPR2020_TPN,feichtenhofer2020x3d}.

\subsection{Vision Transformer}
The transformers \cite{vaswani2017attention} was previously proposed for NLP tasks \cite{devlin2019bert} and recently adopted for computer vision tasks. The transformers were roughly used in three different ways in previous works:
1.To bridge the gap between different modalities, e.g. video captioning \cite{zhou2018end}, video retrieval \cite{gabeur2020multi} and dialog system \cite{li2020bridging}.
2. To aggregate convolutional features for down-streaming tasks, e.g. object detection \cite{dai2020up,carion2020end}, pose estimation \cite{yang2020transpose}, semantic segmentation \cite{duke2021sstvos} and action recognition \cite{girdhar2019video}.
3. To perform feature learning on raw pixels, e.g. most recently image classification \cite{dosovitskiy2020image, touvron2020training}.

Action recognition with self-attention on convolution features \cite{girdhar2019video} is proved successful, however, convolution also generates local feature and gives redundant computations. Different from \cite{girdhar2019video} and inspired by very recent work on applying transformer on raw pixels \cite{dosovitskiy2020image, touvron2020training}, we pioneer the work on aggregating spatio-temporal feature from raw videos without relying on convolution features. 
Different from very recent work \cite{neimark2021video} that extract spatial feature with vision transformer on every video frames and then aggregate feature with attention, our proposed method jointly learns spatio-temporal feature with lower computational cost and higher performance. 
Our work differs from the concurrent work \cite{bertasius2021space}, we present a split attention with better performance without requiring larger video resolution nor extra long clip length. 
Some more recent work \cite{bertasius2021space,fan2021multiscale,bertasius2021space,patrick2021keeping,arnab2021vivit,liu2021video} further studied the multi-scale and different attention factorization methods. 

\section{Video Transformer}
We introduce the Video Transformer starting with the vanilla video transformer (section \ref{sec:v_vidtr}) which illustrates our idea of video action recognition without convolutions. We then present VidTr by first introducing separable-attention (section \ref{sec:s_vidtr}), and then the attention pooling to drop non-representative information temporally (section \ref{sec:s_vidtr}). 

\begin{figure}[t]
	\begin{center}
		\includegraphics[width=0.9\columnwidth]{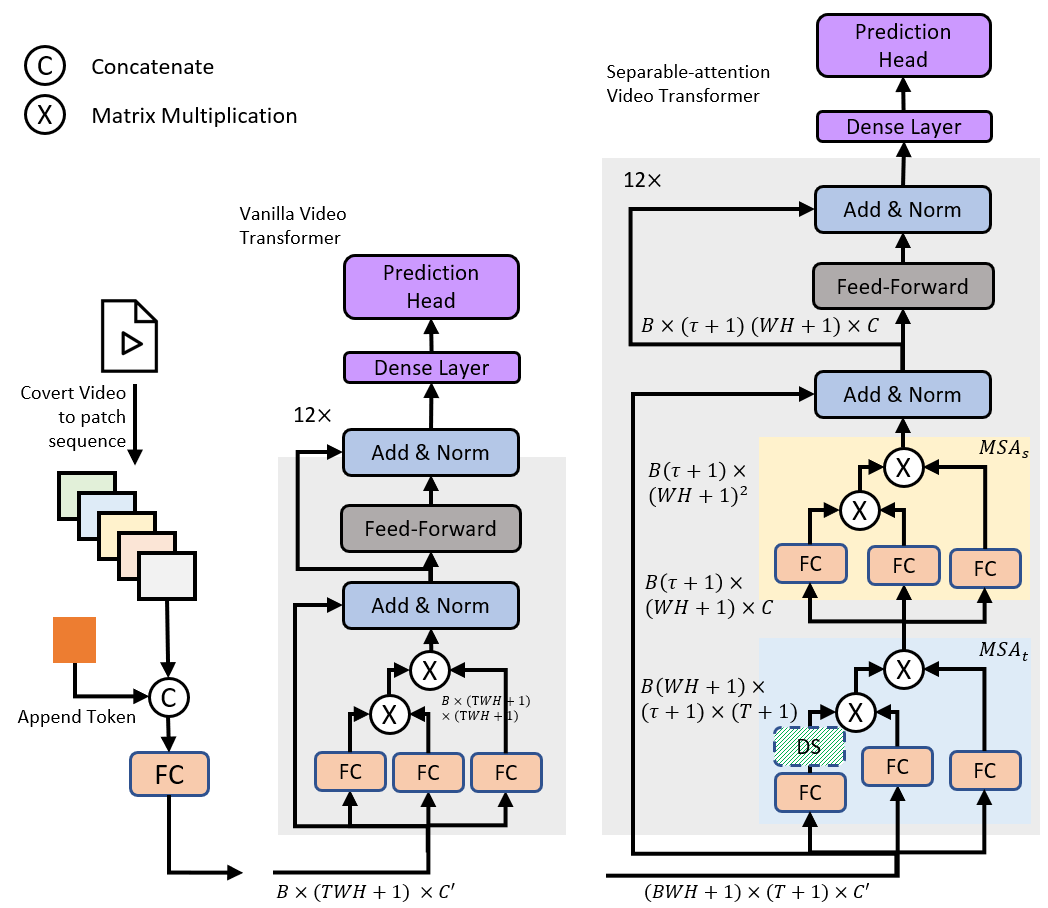}
	\end{center}
	\caption{
		Spatio-temporal separable-attention video transformer (VidTr). The model takes pixels patches as input and learns the spatial temporal feature via proposed separable-attention. The green shaded block denotes the down-sample module which can be inserted into VidTr for higher efficiency. $\tau$ denotes the  temporal dimension after downsampling.
	}
	\label{fig:2}
\end{figure}

\subsection{Vanilla Video Transformer}
\label{sec:v_vidtr}
Following previous efforts in NLP \cite{devlin2019bert} and image classification \cite{dosovitskiy2020image}, we adopt the transformer \cite{vaswani2017attention} encoder structure for action recognition that operates on raw pixels. 
Given a video clip $V \in \mathbb{R}^{C \times T \times W \times H}$, where $T$ denotes the clip length, $W$ and $H$ denote the video frame width and height, and $C$ denotes the number of channel, we first convert $V$ to a sequence of $s\times s$ spatial patches, and apply a linear embedding to each patch, namely $S\in \mathbb{R}^{T\frac{H}{s} \frac{W}{s} \times C^{'}}$, where $C^{'}$ is the channel dimension after the linear embedding.
We add a 1D learnable positional embedding \cite{devlin2019bert, dosovitskiy2020image} to $S$ and following previous work \cite{devlin2019bert, dosovitskiy2020image}, append a class token as well, whose purpose is to aggregate features from the whole sequence for classification. This results in $S'\in \mathbb{R}^{(\frac{TWH}{s^2} + 1) \times C'}$, where $S^{'}_0 \in \mathbb{R}^{1 \times C'}$ is the attached class token. $S'$ is feed into our transformer encoder structure detailed next.

As Figure \ref{fig:2} middle shows, we expand the previous successful ViT transformer architecture for 3D feature learning. Specifically, we stack 12 encoder layers, with each encoder layer consisting of an 8-head self-attention layer and two dense layers with 768 and 3072 hidden units. Different from transformers for 2D images, each attention layer learns a spatio-temporal affinity map $Attn \in \mathbb{R}^{(\frac{TWH}{s^2} + 1) \times (\frac{TWH}{s^2} + 1)}$. 

\subsection{VidTr}
\label{sec:s_vidtr}
In Table \ref{tab:k400_res} we show that this simple formulation is capable of learning 3D motion features on a sequence of local patches. However, as explained in \cite{beltagy2020longformer}, the affinity attention matrix $Attn \in \mathbb{R}^{(\frac{TWH}{s^2} + 1) \times (\frac{TWH}{s^2} + 1)}$ needs to be stored in memory for back propagating, and thus the memory consumption is quadratically related to the sequence length. We can see that the vanilla video transformer increases memory usage for the affinity map from $\mathbb{O}(W^2H^2)$ to $\mathbb{O}(T^2W^2H^2)$, leading to $T^2 \times$ memory usage for training, which makes it impractical on most available GPU devices. We now address this inefficiency with a separable attention architecture.

\subsubsection{Separable-Attention}
To address these memory constraints, we introduce a multi-head separable-attention (MSA) by decoupling the 3D self-attention to a spatial attention $\operatorname{MSA}_s$ and a temporal attention $\operatorname{MSA}_t$ (Figure \ref{fig:2}): 
\begin{equation}
    \operatorname{MSA}(S) = \operatorname{MSA}_s(\operatorname{MSA}_t(S))
\end{equation}

Different from the vanilla video transformer that applies 1D sequential modeling on $S$, we decouple $S$ to a 2D sequence $\hat{S} \in \mathbb{R}^{(T+1) \times (\frac{WH}{s^2}+1) \times C'} $ with positional embedding and two types of class tokens that append additional tokens along the spatial and temporal dimensions. 
Here, the spatial class tokens gather information from spatial patches in a single frame using spatial attention, and the temporal class tokens gather information from patches across frames (at same location) using temporal attention. Then the intersection of the spatial and temporal class tokens $\hat{S}^{(0, 0,:)}$ is used for the final classification. 
To decouple 1D self-attention functions on 2D sequential features $\hat{S}$, we first operate on each spatial location ($i$) independently, applying temporal attention as:
\begin{flalign}
        \hat{S}_t^{(:, i, :)} & = \operatorname{MSA_t}(k=q=v=\hat{S}^{(:, i, :)}) \\
        & = pool(Attn_t) \cdot v_t \\
        & = pool(\operatorname{Softmax}(q_t \cdot k_t^T)) \cdot v_t
\end{flalign}
where $\hat{S}_t \in \mathbb{R}^{(\tau + 1) \times (\frac{WH}{s^2} + 1) \times C}$ is the output of $\operatorname{MSA_t}$,  =
$pool$ denotes the down-sampling method to reduce temporal dimension (from $T$ to $\tau$, $\tau=T$ when no down-sampling is performed) that will be described later, $q_t$, $k_t$, and $v_t$ denote key, query, and value features after applying independent linear functions (LN) on $\hat{S}$:
\begin{equation}
        \footnotesize
        q_t= \operatorname{LN}_q(\hat{S}^{(:, i, :)}); k_t = \operatorname{LN}_k(\hat{S}^{(:, i, :)}); v_t = \operatorname{LN}_v(\hat{S}^{(:, i, :)})
\end{equation}
Moreover, $Attn_t \in \mathbb{R}^{(\tau + 1) \times (T + 1)}$ represent a temporal attention obtained from matrix multiplication between $q_t$ and $k_t$. 
Following $\operatorname{MSA_s}$, we apply a similar 1D sequential self-attention $\operatorname{MSA_s}$ on spatial dimension: 
\begin{flalign}
        \hat{S}_{st}^{(i, :, :)} & = \operatorname{MSA_s}(k=q=v=\hat{S}_t^{(i, :, :)}) \\
        & = Attn_s \cdot v_s \\
        & = \operatorname{Softmax}(q_s \cdot k_s^T) \cdot v_s
\end{flalign}
where $\hat{S}_{st} \in \mathbb{R}^{(\tau + 1) \times (\frac{WH}{s^2} + 1) \times C}$ is the output of $\operatorname{MSA_s}$, $q_s$, $k_s$, and $v_s$ denotes key, query, and value features after applying independent linear functions on $\hat{S}_t$. $Attn_s\in \mathbb{R}^{(\frac{WH}{s^2} + 1) \times (\frac{WH}{s^2} + 1)}$ represent a spatial-wise affinity map. We do not apply down-sampling on the spatial attention as we saw a significant performance drop in our preliminary experiments.

Our spatio-temporal split attention decreased the memory usage of the transformer layer by reducing the affinity matrix from $\mathbb{O}(T^2W^2H^2)$ to $\mathbb{O}(\tau^2 + W^2H^2)$. This allows us to explore longer temporal sequence lengths that were infeasible on modern hardware with the vanilla transformer.

\subsubsection{Temporal Down-sampling method}
\label{sec:c_vidtr}
Video content usually contains redundant information \cite{li2020nuta}, with multiple frames depicting near identical content over time. We introduce compact VidTr (C-VidTr) by applying temporal down-sampling within our transformer architecture to remove some of this redundancy. 
We study different temporal down-sampling methods ($pool$ in Eq. 3) including temporal average pooling and 1D convolutions with stride 2, which reduce the temporal dimension by half (details in Table \ref{tab:ablation_temp_ds}).  

A limitation of these pooling the methods is that they uniformly aggregate information across time but often in video clips the informative frames are not uniformly distributed. We adopted the idea of non-uniform temporal feature aggregation from previous work \cite{li2020nuta}.
Different from previous work \cite{li2020nuta} that directly down-sample the query using average pooling, we found that in our proposed network, the temporal attention highly activates on a small set of temporal features when the clip is informative, while the attention equally distributed over the length of the clip when the clip caries little additional semantic information.
Building on this intuition, we propose a topK based pooling ($topK\_std$ pooling) that orders instances by the standard deviation of each row in the attention matrix: 
\begin{equation}
\small
    \operatorname{pool}_{topK\_std}(Attn_t^{(1:,:)}) = Attn_t^{\left(\operatorname{topK}\left(\sigma(Attn_t^{(1:,:)})\right), : \right)}
\end{equation}
where $\sigma \in \mathbb{R}^{T}$ is row-wise standard deviation of $Attn_t^{(1:,:)}$ as:
\begin{flalign}
\small
    \sigma^{(i)} & = \frac{1}{T}\sqrt{ \sum_{i=1}^{T} (Attn_t^{(i, :)} - \mu)^2} \\
    \mu^{(i)} &  = \frac{1}{T}\sum_{i=1}^{T} Attn_t^{(i, :)}
\end{flalign}
where $\mu \in \mathbb{R}^{T}$ is the mean of $Attn_t^{(1:,:)}$. Note that the $topK\_std$ pooling was applied to the affinity matrix excludes the token ($Attn_t^{(0,:,:)}$) as we always preserve token for information aggregation. 
Our experiments show that $topK\_std$ pooling gives better performance than average pooling or convolution. The $topK\_std$ pooling can be intuitively understood as  selecting the frames with strong localized attention and removing frames with uniform attention.

\subsection{Implementation Details} 
\begin{table}[t!]
    \footnotesize
	\begin{center}
		\begin{tabularx}{1\columnwidth}{l|c|c|l|l} 
		    \toprule
			Model        & $clip\_len$ & $sr$ & Down-sample Layer & $\tau$     \\ 
			\midrule
			VidTr-S        & 8 & 8 & - & -  \\
			VidTr-M        & 16 & 4 & - & -  \\
			VidTr-L        & 32 & 2 & - & -   \\
			C-VidTr-S      & 8 & 8 & [1,2,4] & [6,4,2]   \\
			C-VidTr-M      & 16 & 4 & [1,2,4] & [8,4,2] \\
			\bottomrule
		\end{tabularx}
	\end{center}
	\caption{Detailed configuration of different VidTr permutations. $clip\_len$ denotes the sampled clip length and $sr$ stands for the sample rate. We uniformly sample $clip\_len$ frames out of $clip\_len \times sr$ consecutive frames. The configurations are empirically selected, details in \nameref{tab:ablation}.}
	\label{tab:permutation}
\end{table}

\textbf{Model Instantiating:} Based on the input clip length and sample rate, we introduce three base VidTr models (VidTr-S,VidTr-M and VidTr-L). By applying the different pooling strategies we introduce two compact VidTr permutations (C-VidTr-S, and C-VidTr-M). 
To normalize the feature space, we apply layer normalization before and after the residual connection of each transformer layer and adopt the GELU activation as suggested in \cite{dosovitskiy2020image}. Detailed configurations can be found in Table \ref{tab:permutation}. We empirically determined the configuration for different clip length to produce a set of models from low FLOPs and low latency to high 
accuracy (details in \nameref{tab:ablation}). 

During \textbf{training} we initialize our model weights from ViT-B \cite{dosovitskiy2020image}. To avoid over fitting, we adopted the commonly used augmentation strategies including random crop, random horizontal flip (except for Something-something dataset). We trained the model using 64 Tesla V100 GPUs, with batch size of 6 per-GPU (for VidTr-S) and weight decay of 1e-5.
We adopted SGD as the optimizer but found the Adam optimizer also gives us the same performance. We trained our network for 50 epochs in total with initial learning rate of 0.01, and reduced it by 10 times after epochs 25 and 40. It takes about 12 hours for VidTr-S model to converge, the training process also scales well with fewer GPUs (e.g. 8 GPUs for 4 days).
During \textbf{inference} we adopted the commonly used 30-crop evaluation for VidTr and compact VidTr, with 10 uniformly sampled temporal segments and 3 uniformly sampled spatial crop on each temporal segment \cite{wang2018non}. It is worth mentioning that we can further boost the inference speed of compact VidTr by adopting a single pass inference mechanise, this is because the attention mechanism captures global information more effectively than 3D convolution. We do this by training a model with frames sampled in TSN \cite{wang_ECCV2016_TSN} style, and uniformly sampling $N$ frames in inference (details in supplemental materials).

\section{Experimental Results}
\subsection{Datasets}
We evaluate our method on six of the most widely used datasets.
\textbf{Kinetics 400~\cite{kay2017kinetics} and Kinetics 700~\cite{carreira2019short}} consists of approximately 240K/650K training videos and 20K/35K validation videos trimmed to 10 seconds from 400/700 human action categories. We report top-1 and top-5 classification accuracy on the validation sets.
\textbf{Something-Something V2~\cite{goyal2017something}} dataset consists of 174 actions and contains 168.9K training videos and 24.7K evaluation videos. We report top-1 accuracy following previous works~\cite{lin2019tsm} evaluation setup. 
\textbf{Charades~\cite{sigurdsson2016hollywood}} has 9.8k training videos and 1.8k validation videos spanning about 30 seconds on average. Charades contains 157 multi-label classes with longer activities, performance is measured in mean Average Precision (mAP).
\textbf{UCF-101~\cite{soomro2012dataset} and HMDB-51~\cite{kuehne2011hmdb}} are two smaller datasets. UCF-101 contains 13320 videos with an average length of 180 frames per video and 101 action categories. The HMDB-51 has 6,766 videos and 51 action categories. We report the top-1 classification on the validation videos based on split 1 for both dataset.

\subsection{Kinetics 400 Results}
\subsubsection{Comparison To SOTA} 

\begin{table}[h!]
    \footnotesize
	\begin{center}
	\scalebox{0.97}{
		\begin{tabularx}{1.03\columnwidth}{l|c|c|c|c|c} 
		    \toprule
			Model        & Input & GFLOPs & Lat. & top-1 &  top-5  \\ 
			\midrule
			I3D50~\cite{yang2020temporal}   & $32\times2$   & 167 & 74.4 & 75.0 & 92.2 \\
			I3D101~\cite{yang2020temporal}  & $32\times2$   & 342 & 118.3 & 77.4 & 92.7 \\
			NL50~\cite{wang2018non}    & $32\times2$   &  282& 53.3 & 76.5  & 92.6 \\
			NL101~\cite{wang2018non}   & $32\times2$   & 544 & 134.1& 77.7 & 93.3 \\
			TEA50~\cite{li2020tea}   & $16\times2$   & 70 &  - & 76.1 & 92.5 \\
			TEINet~\cite{liu_AAAI2020_TEINet}   & $16\times2$   & 66 &  49.5 & 76.2 & 92.5 \\
			CIDC~\cite{li2020directional}   & $32\times2$   & 101 & 82.3  & 75.5 & 92.1 \\
			SF50~8$\times$8~\cite{feichtenhofer2018slowfast}     & (32+8)$\times2$   & 66 & 49.3 & 77.0 & 92.6  \\
			SF101~8$\times$8~\cite{feichtenhofer2018slowfast}   & (32+8)$\times2$   & 106 & 71.9 & 77.5 & 92.3  \\
			SF101~16$\times$8~\cite{feichtenhofer2018slowfast}   & (64+16)$\times2$   & 213 & 124.3 & 78.9 & 93.5  \\
			TPN50~\cite{yang2020temporal}   & $32\times2$   & 199 & 89.3 & 77.7 & 93.3 \\
			TPN101~\cite{yang2020temporal}   & $32\times2$   & 374 & 133.4 & 78.9 & 93.9 \\  
			CorrNet50~\cite{wang2020video}   & $32\times2$   & 115 & - & 77.2 & N/A \\
			CorrNet101~\cite{wang2020video}   & $32\times2$   & 187 & - & 78.5 & N/A \\
			X3D-XXL~\cite{feichtenhofer2020x3d}   & $16\times5$   & 196 & - & 80.4 & 94.6 \\
			\midrule
			Vanilla-Tr & $8\times8$ & 89 & 32.8 & 77.5 & 93.2 \\
			VidTr-S              & $8\times8$ & 89 & 36.2 & 77.7 & 93.3 \\
			VidTr-M              & $16\times4$ & 179 & 61.1 & 78.6 & 93.5  \\
			VidTr-L              & $32\times2$ & 351 & 110.2 & 79.1 & 93.9 \\
			\midrule
			En-I3D-50-101              & $32\times2$ & 509 & 192.7 & 77.7 &  93.2\\
			En-I3D-TPN-101  & $32\times2$ & 541 & 207.8 & 79.1 &  94.0\\
			\midrule
			En-VidTr-S              & $8\times8$ & 130 & 73.2 & 79.4 &  94.0\\
			En-VidTr-M              & $16\times4$ & 220 & 98.1 & 79.7 &  94.2\\
			En-VidTr-L              & $32\times2$ & 392 & 147.2 &  80.5 &  94.6\\
			\bottomrule
		\end{tabularx}}
	\end{center}
	\caption{Results on Kinetics-400 dataset. We report top-1 accuracy(\%) on the validation set. The `Input' column indicates what frames of the 64 frame clip are actually sent to the network. $n \times s$ input indicates we feed $n$ frames to the network sampled every $s$ frames. Lat. stands for the latency on single crop.}
	\label{tab:k400_res}
\end{table}

We report results on the validation set of Kinetics 400 in Table \ref{tab:k400_res}, including the top-1 and top-5 accuracy, GFLOPs (Giga Floating-Point Operations) and latency (ms) required to compute results on one view. 

As shown in Table \ref{tab:k400_res}, the VidTr achieved the SOTA performance compared to previous I3D based SOTA architectures with lower GFLOPs and latency.
The VidTr significantly outperform previous SOTA methods at roughly same computational budget, e.g. at ~200 GFLOPs, the VidTr-M outperform I3D50 by $3.6\%$, NL50 by $2.1\%$,and TPN50 by $0.9\%$.
At similar accuracy levels, VidTr is significantly more computationally efficient than other works, e.g. at  ~$78\%$ top-1 accuracy, the VidTr-S has 6$\times$ fewer FLOPs than NL-101, 2$\times$ fewer FLOPs than TPN and $12\%$ fewer FLOPs than Slowfast-101.
We also see that our VidTr outperforms I3D based networks at higher sample rate (e.g. $s =8$, TPN achieved $76.1\%$ top-1 accuracy), this denotes, the global attention learns temporal information more effectively than 3D convolutions.
X3D-XXL from architecture search is the only network that outperforms our VidTr. We plan to use architecture search techniques for attention based architecture in future work.

\subsubsection{Compact VidTr}
We evaluate the effectiveness of our compact VidTr with the proposed temporal down-sampling method (Table \ref{tab:permutation}). 
The results (Table \ref{tab:compact_VidTr}) show that the proposed down-sampling strategy removes roughly $56\%$ of the computation required by VidTr with only $2\%$ performance drop in accuracy. The compact VidTr complete the VidTr family from small models (only 39GFLOPs) to high performance models (up to 79.1\% accuracy).
Compared with previous SOTA compact models~\cite{li2020tea,liu_AAAI2020_TEINet}, our compact VidTr achieves better or similar performance with lower FLOPs and latency, including: TEA (+0.6\% with 16\% fewer FLOPs) and TEINet (+0.5\% with 11\% fewer FLOPs).
\begin{table}[t]
    \footnotesize
	\begin{center}
	\scalebox{0.95}{
		\begin{tabularx}{1.05\columnwidth}{l|c|c|c|c|c} 
		    \toprule
			Model        &Input & Res.& GFLOPs & Latency(ms) &  top-1  \\ 
			\midrule
			TSM~\cite{lin2019tsm} & $8f$TSN & 256  & 69   & $29$ & 74.7  \\
			TEA~\cite{li2020tea} & 16$\times$4  & 256  & 70   & - & 76.1  \\
			3DEffi-B4~\cite{feichtenhofer2020x3d}& 16$\times$5 & 224  & 7   & - & 72.4  \\
            TEINet~\cite{liu_AAAI2020_TEINet} & 16$\times$4 & 256 & 33 & $36$ & 74.9 \\
			X3D-M~\cite{feichtenhofer2020x3d} & 16$\times$5  & 224  & 5   & $40.9$ & 74.6  \\
			X3D-L~\cite{feichtenhofer2020x3d}   & 16$\times$5  & 312 & 19   & $59.4$ & 76.8  \\
			\midrule
			C-VidTr-S                & 8$\times$8 & 224 & 39 & $17.5$ & 75.7 \\
			C-VidTr-M              & 16$\times$4 & 224 & 59 & $26.1$ & 76.7\\
			\bottomrule
		\end{tabularx}}
	\end{center}
	\caption[Caption for LOF]{Comparison of VidTr to other fast networks. We present the number of views used for evaluation and FLOPs required for each view. The latency denotes the total time required to get the reported top-1 score.\protect\footnotemark }
	\label{tab:compact_VidTr}
\end{table}

\subsubsection{Error and Ensemble Analysis}
\begin{table}[t!]
\footnotesize
    \scalebox{0.95}{
    \subfloat[Top 5 classes that VidTr works better than I3D.]{
		\begin{tabularx}{0.45\columnwidth}{l|c} 
			\toprule
			Top 5 (+)        & Acc. gain   \\ 
			\midrule
			 	making a cake     &   +26.0\%     \\ 
		        catching fish &  +21.2\%  \\ 
			    catching baseball    &   +20.8\%    \\ 
			    stretching arm & +19.1\%   \\ 
			    spraying & + 18.0 \%\\
			\bottomrule
		\end{tabularx}
    }}
    \scalebox{0.95}{
    \subfloat[Top 5 classes that I3D works better than VidTr.]{
	\begin{tabularx}{0.45\columnwidth}{l|c} 
		\toprule
		Top 5 (-) & Acc. gain   \\ 
		\midrule
		 	shaking head & -21.7\%     \\ 
	        dunking basketball &    -20.8\%  \\ 
		    lunge  & -19.9\%   \\ 
		    playing guitar &   -19.9\%  \\ 
		    tap dancing & -16.3\%\\
		\bottomrule
	\end{tabularx}
	}
    } \hfill
	\caption{ Quantitative analysis on Kinetics-400 dataset. The performance gain is defined as the disparity of the top-1 accuracy between VidTr network and that of I3D.}
	\label{tab:supp_error}
\end{table}
We compare the errors made by VidTr-S and the I3D50 network to better understand the local networks' (I3D) and global networks' (VidTr) behavior. We provide the top-5 activities that our VidTr-S gain most significant improvement over the I3D50. We find that our VidTr-S outperformed the I3D on the activities that requires long-term video contexts to be recognized. For example, our VidTr-S outperformed the I3D50 on ``making a cake'' by 26\% in accuracy. The I3D50 overfits to ``cakes'' and often recognize making a cake as eating a cake. We also analyze the top-5 activities where I3D does better than our VidTr-S (Table \ref{tab:supp_error}). Our VidTr-S performs poorly on the activities that need to capture fast and local motions. For example, our VidTr-S performs 21\% worse in accuracy on ``shaking head''. 

Inspired by the findings in our error analysis, we ensembled our VidTr with a light weight I3D50 network by averaging the output values between the two networks. The results (Table \ref{tab:k400_res}) show that the the I3D model and transformer model complements each other and the ensemble model roughly lead to 2\% performance improvement on Kinetics 400 with limited additional FLOPs (37G). The performance gained by ensembling the VidTr with I3D is significantly better than the improvement by combine two 3D networks (Table \ref{tab:k400_res}).

\footnotetext{we measure latency of X3D using the authors' code and fast depth convolution patch: \url{https://github.com/facebookresearch/SlowFast/blob/master/projects/x3d/README.md}, which only has models for X3D-M and X3D-L and not the XL and XXL variants}

\subsubsection{Ablations}
\begin{table}[h]
    \footnotesize
	\centering
	\subfloat[Comparison between different patching strategies.]{
	\scalebox{0.95}{
		\begin{tabularx}{0.23\textwidth}{l|c|c} 
			\toprule
			Model        & FP. & top-1   \\ 
			\midrule
			Cubic (4$\times$16$^2$)   &  23G  & 73.1    \\ 
			Cubic (2$\times$16$^2$)   &  45G  & 75.5    \\ 
			Square (1$\times$16$^2$)  &  89G  & 77.7  \\ 
			Square (1$\times$32$^2$)   &  21G  & 71.2 \\
			\bottomrule
		\end{tabularx}}
		\label{tab:ablation_patch}
	} \hfill
    \subfloat[Comparison between different factorization. ]{
    \scalebox{0.95}{
    	\begin{tabularx}{0.23\textwidth}{l|c|c} 
    		\toprule
    		Model       &Mem.   & top-1   \\ 
    		\midrule
    		WH & 2.1GB &  74.7  \\ 
    		WHT & 7.6GB &  77.5\\ 
    		\textbf{WH + T} & 2.3GB & 77.7 \\
    		W + H + T. & 1.5GB & 72.3 \\
    		\bottomrule
    	\end{tabularx}
		\label{tab:ablation_temp_modeling}
    }}

    \subfloat[Comparison between different backbones.]{
    \scalebox{0.95}{
	\begin{tabularx}{0.2\textwidth}{l|c|c} 
		\toprule
		Init. from & FP.&top-1   \\ 
		\midrule
		T2T~\cite{yuan2021tokens} & 34G  & 76.3 \\ 
		ViT-B~\cite{dosovitskiy2020image} & 89G  & 77.7  \\ 
		ViT-L~\cite{dosovitskiy2020image} & 358 & 77.5 \\
		\bottomrule
	\end{tabularx}}
	\label{tab:ablation_backbone}
    }\hfill
    \subfloat[Comparison between different down-sample methods.]{
    \scalebox{0.95}{
	\begin{tabularx}{0.25\textwidth}{l|c|c} 
		\toprule
		Configurations & top-1        & top-5   \\ 
		\midrule
		Temp. Avg. Pool.  &  74.9 & 91.6  \\ 
		1D Conv.~\cite{yuan2021tokens}    &   75.4 & 92.3    \\ 
		STD Pool.         &  75.7&   92.2 \\
		\bottomrule
	\end{tabularx}}
	\label{tab:ablation_temp_ds}
    }\hfill
    \subfloat[Compact VidTr down-sampling twice at layer $k$ and $k+2$.]{
    \scalebox{0.95}{
    	 \begin{tabularx}{0.47\columnwidth}{l|c|c|c} 
    		\toprule
    		Layer & $\tau$ & FP.& top-1   \\ 
    		\midrule
    		$[0,2]$ & $[4,2]$&  26G & 72.9 \\ 
    		$[1,3]$ & $[4,2]$& 32G  & 74.9  \\ 
    		$[2,4]$ & $[4,2]$& 47G & 74.9 \\
    		$[6,8]$ & $[4,2]$& 60G & 75.3 \\
    		\bottomrule
    	\end{tabularx}}
	\label{tab:ablation_where_a}
    }\hfill
    \subfloat[Compact VidTr down-sampling twice starting from layer $1$ and skipping different number of layers.]{
    \scalebox{0.95}{
    	 \begin{tabularx}{0.47\columnwidth}{l|c|c|c} 
    		\toprule
    		Layer & $\tau$ & FP.& top-1   \\ 
    		\midrule
    		$[1,2]$ & $[4,2]$&  30G & 73.9 \\ 
    		$[1,3]$ & $[4,2]$& 32G  & 74.9  \\ 
    		$[1,4]$ & $[4,2]$& 33G & 75.0 \\
    		$[1,5]$ & $[4,2]$& 34G & 75.2 \\
    		\bottomrule
    	\end{tabularx}}
	\label{tab:ablation_where_b}
    }
	\label{tab:ablation}
    \caption{ Ablation studies on Kinetics 400 dataset. 
    We use an VidTr-S backbone with 8 frames input for (a,b) and C-VidTr-S for (c,d).
	The evaluation is performed on 30 views with 8 frame input unless specified. FP. stands for FLOPs.} 
\end{table}
We perform all ablation experiments with our VidTr-S model on Kinetics 400. We used $8 \times 224 \times 224$ input with a frame sample rate of 8, and 30-view evaluation.\\
\textbf{Patching strategies: } 
We first compare the cubic patch ($4\times 16^2$), where the video is represented as a sequence of spatio-temporal patches, with the square patch ($1\times16^2$), where the video is represented as a sequence of spatial patches. Our results (Table \ref{tab:ablation_patch}) show that the model using cubic patches with longer temporal size has fewer FLOPs but results in significant performance drop (73.1 vs. 75.5). The model using square patches significantly outperform all cubic patch based models, likely because the linear embedding is not enough to represent the shot-term temporal association in the cubic. We further compared the performance of using different patch sizes ($1\times16^2$ vs. $1\times32^2$), using $32^2$ patches lead to $4\times$ decreasing of the sequence length, which decreases memory consumption of the affinity matrices by $16\times$, however, using $16^2$ patches significantly outperform the model using $32^2$ patches (77.7 vs. 71.2). We did not evaluate the model using smaller patching sizes (e.g., $8 \times 8$) because of the high memory consumption.
\\
\textbf{Attention Factorization:}
We compare different factorization for attention design, including spatial modeling only (WH), jointly spatio-temporal modeling module (WHT, vanilla-Tr), spatio-temporal separable-attention (WH + T, VidTr), and axial separable-attention (W + H + T).
We first evaluate an spatio-only transformer. We average the class token for each input frame for our final output. Our results (Table \ref{tab:ablation_temp_modeling}) show that the spatio-only transformer requires less memory but has worse performance compare with spatio-temporal attention models. This shows that temporal modeling is critical for attention based architectures. The joint spatio-temporal transformer significantly outperforms the spatio-only transformer but requires a restrictive amount of memory ($T^2$ times for the affinity matrices). Our VidTr using spatio-temporal separable-attention requires $3.3\times$ less memory with no accuracy drop. We further evaluate the axial separable-attention (W + H + T), which requires the least memory. The results (Table \ref{tab:ablation_temp_modeling}) show that the axial separable-attention has a significant performance drop likely due to breaking the X and Y spatial dimensions. 
\\
\textbf{Sequence down-sampling comparison: }
We compare different down-sampling strategy including temporal average pooling, 1D temporal convolution and the proposed STD-based topK pooling method. The results (Table \ref{tab:ablation_temp_ds}) show that our proposed STD-based down-sampling method outperformed the temporal average pooling and the convolution-based down-sampling strategies that uniformly aggregate information over time.
\\
\begin{figure}[t]
\begin{center}
    \includegraphics[width=\columnwidth]{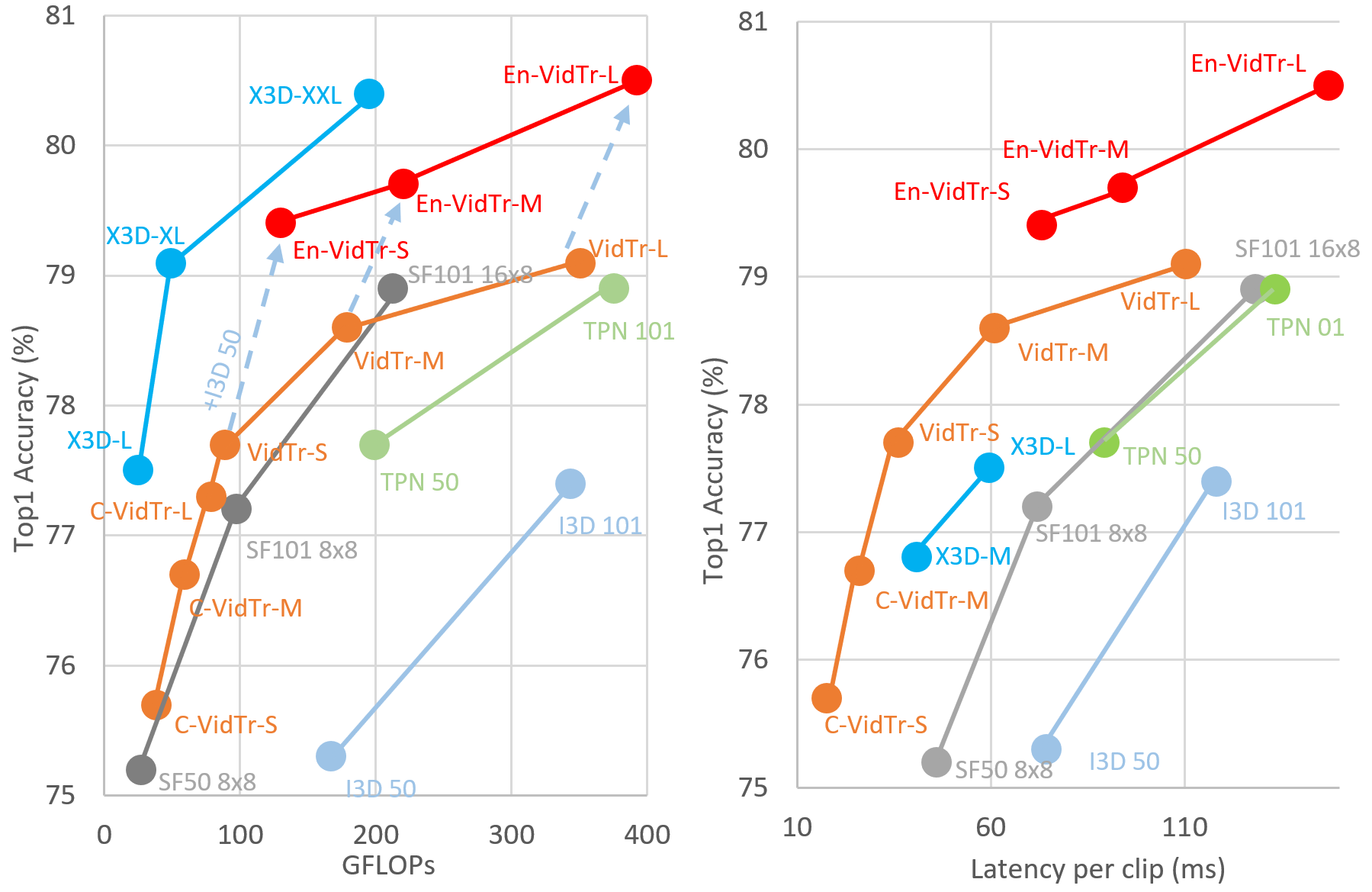}
	\caption{The comparison between different models on accuracy, FLOPs and latency.}
	\label{fig:k400_FLOPs_latency}
\end{center}
\end{figure}
\textbf{Backbone generalization: }
We evaluate our VidTr initialized with different models, including T2T~\cite{yuan2021tokens}, ViT-B, and ViT-L. The results on Table \ref{tab:ablation_backbone} show that our VidTr achieves reasonable performance across all backbones. The VidTr using T2T as the backbone has the lowest FLOPs but also the lowest accuracy. The Vit-L-based VidTr achieve similar performance with the Vit-B-based VidTr even with $3 \times$ FLOPs.  As showed in previous work \cite{dosovitskiy2020image}, transformer-based network are more likely to over-fit and Kinetics-400 is relatively small for Vit-L-based VidTr.
\\
\textbf{Where to down-sample: }
Finally we study where to perform temporal down-sampling. We perform temporal down-sampling at different layers (Table \ref{tab:ablation_where_a}). Our results (Table \ref{tab:ablation_where_a}) show that starting to perform down-sampling after the first encoder layer has the best trade-off between the performance and FLOPs. Starting to perform down-sampling at very beginning leads to the fewest FLOPs but has a significant performance drop (72.9 vs. 74.9). Performing down-sampling later only has slight performance improvement but requires higher FLOPs. We then analyze how many layers to skip between two down-sample layers. Based on the results in Table \ref{tab:ablation_where_b}, skipping one layer between two down-sample operations has the best trade-off. Performing down-sampling on consecutive layers (0 skip layers) has lowest FLOPs but the performance decreases (73.9 vs. 74.9). Skipping more layers did not show significant performance improvement but does have higher FLOPs.

\begin{table}[t!]
\footnotesize
    \begin{center}
        \begin{tabularx}{\columnwidth}{l|c|c|c|c|c|c} 
	    \toprule
		Model        & Input   & K700 &  Chad & SS & UCF & HM\\ 
		\midrule
		I3D~\cite{carreira2017quo} & 32$\times$2 & 58.7  & 32.9& 50.0 &95.1 & 74.3\\
		TSM~\cite{lin2019tsm} & 8(TSN)  & - & -& 59.3 &94.5 & 70.7\\
		I3D101~\cite{yang_CVPR2020_TPN} & 32 $\times$ 4 & 40.3& - &-&- \\
		CSN152~\cite{tran2019video} &  $32\times2$  & 70.1  & - & - & - & - \\ 
        TEINet\cite{liu_AAAI2020_TEINet} & 16 (TSN)  &-& - & 62.1 &96.7 & 73.3\\
        SF101~\cite{feichtenhofer2018slowfast} & 64$\times$2  & 70.2 & -& 60.9&-&-\\
        SF101-NL~\cite{feichtenhofer2018slowfast} & 64$\times$2 & 70.6 & 45.2&-&-&-\\
        X3D-XL~\cite{feichtenhofer2020x3d} & 16 $\times$ 5  & - & 47.1&-&-&-\\
        \midrule
        VidTr-M & 16 $\times$ 4 & 69.5 & - & 61.9 & 96.6 & 74.4\\
		VidTr-L & 32 $\times$ 2  & 70.2&43.5 & 63.0 & \textbf{96.7} & \textbf{74.4}\\
		En-VidTr-L & 32 $\times$ 2  & 70.8& \textbf{47.3} & - & - & - \\
		\bottomrule
	\end{tabularx}
    \end{center}
	\caption{Results on Kinetics-700 dataset (K700), Charades dataset (Chad), something-something-V2 dataset (SS), UCF-101 and HMDB (HM) dataset. The evaluation metrics are mean average precision (mAP) in percentage for Charades (32$\times$4 input is used), top-1 accuracy for Kinetics 700, something-something-V2 (TSN styled dataloader is used), UCF and HMDB.} 
	\label{tab:charades_res}
\end{table}

\subsubsection{Run-time Analysis}
We further analyzed the trade-off between latency, FLOPs and accuracy.
We note that the VidTr achieved the best balance between these factors (Figure \ref{fig:k400_FLOPs_latency}). The VidTr-S achieve similar performance but significantly fewer FLOPs compare with I3D101-NL ($5\times$ fewer FLOPs), Slowfast101 $8\times8$ ($12\%$ fewer FLOPs), TPN101 ($2\times$ fewer FLOPs), and CorrNet50 ($20\times$ fewer FLOPs). Note that the X3D has very low FLOPs but high latency due to the use of depth convolution. Our experiments show that the X3D-L has about $3.6 \times$ higher latency comparing with VidTr-S (Figure \ref{fig:k400_FLOPs_latency}).
\subsection{More Results}
\paragraph{Kinetics-700 Results: }
Our experiments show a consistent performance trend on Kinetics 700 (Table \ref{tab:charades_res}). 
The VidTr-S significantly outperformed the baseline I3D model (+9\%), the VidTr-M achieved the performance comparable to Slowfast101 $8\times8$ and the VidTr-L is comparable to previous SOTA slowfast101-nonlocal.
There is a small performance gap between our model and Slowfast-NL \cite{feichtenhofer2018slowfast}, because Slowfast is pre-trained on both Kinetics 400 and 600 while we only pre-trained on Kinetics 400. 
Previous findings that VidTr and I3D are being complementary is consistent on Kinetics 700, ensemble VidTr-L with I3D leads to +0.6\% performance boost.\\
\textbf{Charades Results: }
We compare our VidTr with previous SOTA models on Charades. 
Our VidTr-L outperformed previous SOTA methods LFB and NUTA101, and achieved the performance comparable to Slowfast101-NL (Table \ref{tab:charades_res}). The results on Charades demonstrates that our VidTr generalizes well to multi-label activity datasets. 
Our VidTr performs worse than the current SOTA networks (X3D-XL) on Charades likely due to overfitting. 
As discussed in previous work~\cite{dosovitskiy2020image}, the transformer-based networks overfit easier than convolution-based models, and Charades is relatively small. 
We observed a similar finding with our ensemble, ensembling our VidTr with a I3D network (40.3 mAP) achieved SOTA performance.\\
\textbf{Something-something V2 Results: }
We observe that the VidTr does not work well on the something-something dataset (Table \ref{tab:charades_res}), likely because pure transformer based approaches do not model local motion as well as convolutions. This aligns with our observation in our error analysis. 
Further improving  local motion modeling ability is an area of future work.\\
\textbf{UCF and HMDB Results: }
Finally we train our VidTr on two small dataset UCF-101 and HMDB-51 to test if VidTr generalizes to smaller datasets. The VidTr achieved SOTA comparable performance with 6 epochs of training (96.6\% on UCF and 74.4\% on HMDB), showing that the model generalize well on small dataset (Table \ref{tab:charades_res}).

\section{Visualization and Understanding VidTr}
\begin{figure}[t]
\begin{center}
    \subfloat[The spatial and temporal attention in Vidtr. The attention is able to focus on the informative frames and regions.]{
	    \includegraphics[width=0.9\columnwidth]{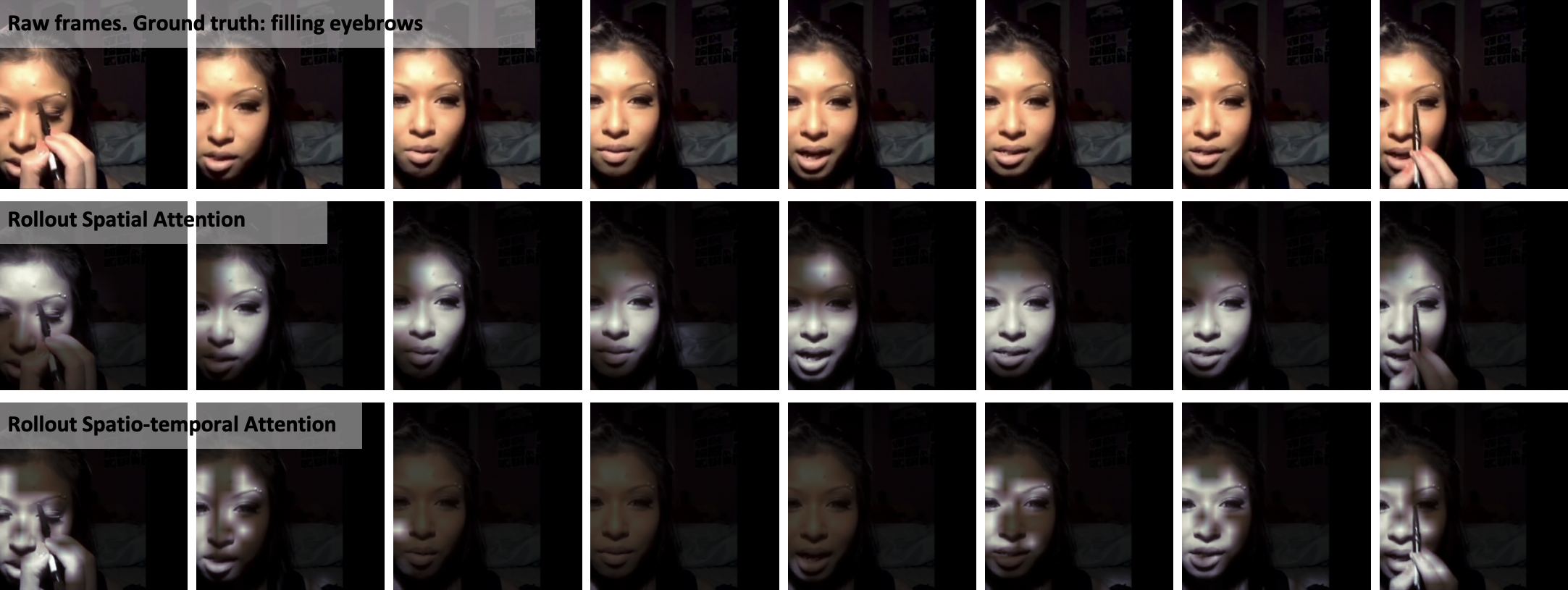}
    \label{fig:vis_a}
	}
	\hfill
    \subfloat[The rollout attentions from different layers of VidTr.]{
	    \includegraphics[width=0.9\columnwidth]{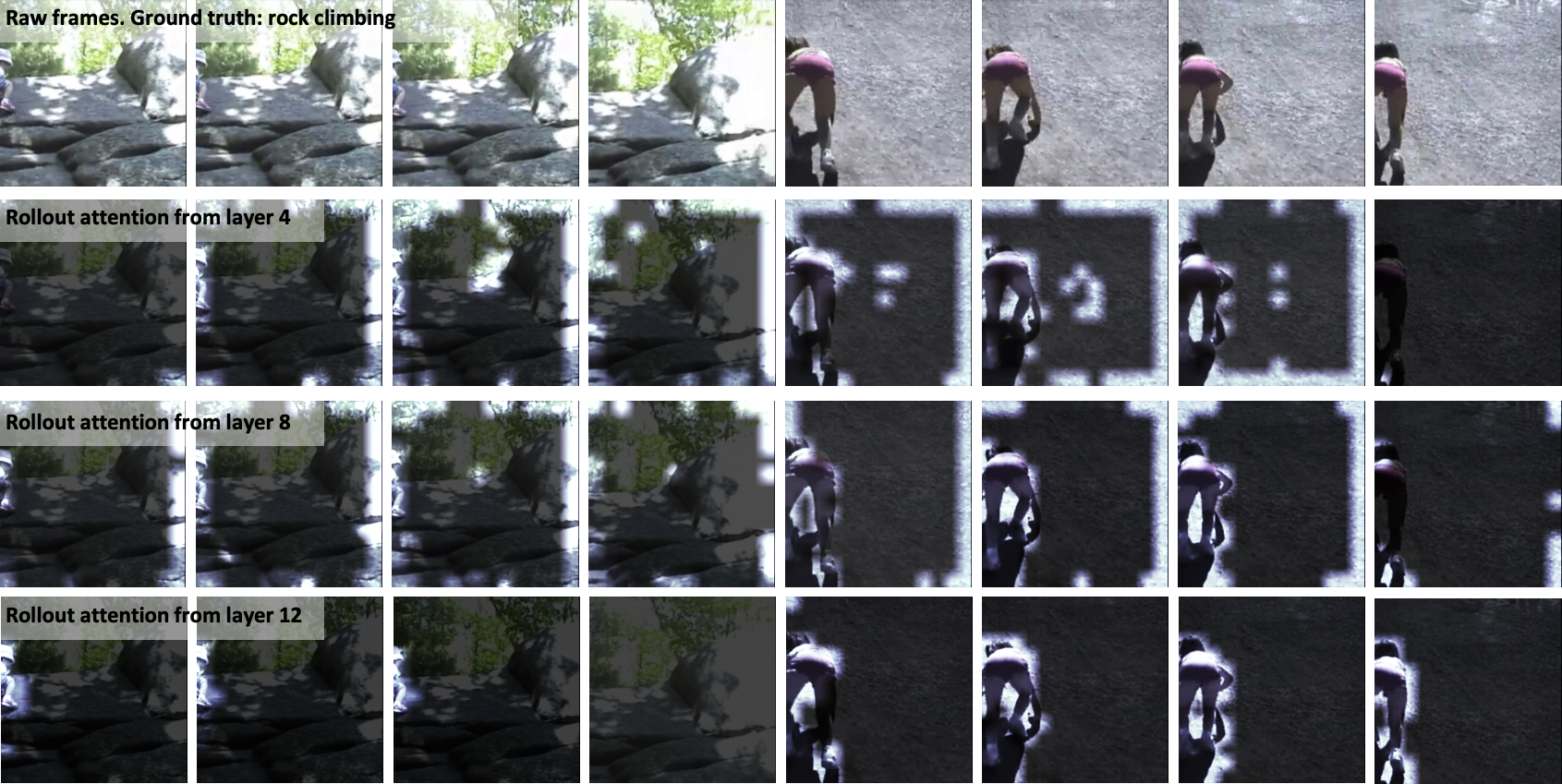}
    \label{fig:vis_b}
	} 
	\hfill 
    \subfloat[Comparison of I3D activations and VidTr attentions.]{
	    \includegraphics[width=0.9\columnwidth]{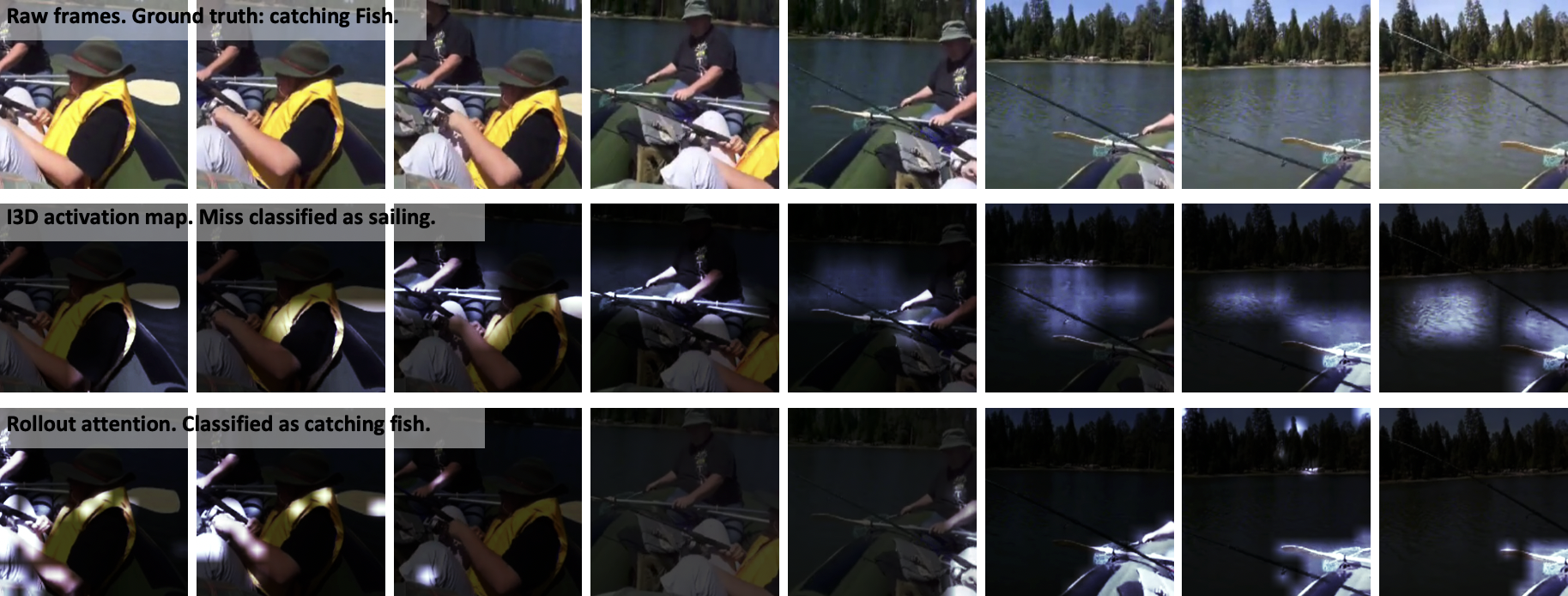}
    \label{fig:vis_c}
	} 
	\caption{Visualization of spatial and temporal attention of VidTr and comparison with I3D activation. 
	}
\end{center}
\end{figure}
We first visualized the VidTr's separable-attention with attention roll-out method \cite{abnar2020quantifying} (Figure \ref{fig:vis_a}). We find that the spatial attention is able to focus on informative regions and temporal attention is able to skip the duplicated/non-representative information temporally.
We then visualized the attention at 4th, 8th and 12th layer of VidTr (Figure \ref{fig:vis_b}), 
we found the spatial attention is stronger on deeper layers.
The attention does not capture meaningful temporal instances at early stages because the temporal feature relies on the spatial information to determine informative temporal instances. 
Finally we compared the I3D activation map and rollout attention from VidTr (Figure \ref{fig:vis_c}). The I3D mis-classified the catching fish as sailing, as the I3D attention focused on the people sitting behind and water. The VidTr is able to make the correct prediction and the attention showed that the VidTr is able to focus on the action related regions across time.

\section{Conclusion}
In this paper, we present video transformer with separable-attention, an novel stacked attention based architecture for video action recognition.
Our experimental results show that the proposed VidTr achieves state-of-the-art or comparable performance on five public action recognition datasets. 
The experiments and error analysis show that the VidTr is especially good at modeling the actions that requires long-term reasoning.
Further combining the advantage of VidTr and convolution for better local-global action modeling \cite{wu2021cvt,liu2021swin} and adopt self-supervised training \cite{chen2021empirical} on large-scaled data will be our future work.\\

{\small
\bibliographystyle{ieee_fullname}
\bibliography{egbib}

\begin{thebibliography}{10}\itemsep=-1pt

\bibitem{abnar2020quantifying}
Samira Abnar and Willem Zuidema.
\newblock Quantifying attention flow in transformers.
\newblock In {\em Proceedings of the 58th Annual Meeting of the Association for
  Computational Linguistics}, pages 4190--4197, 2020.

\bibitem{arnab2021vivit}
Anurag Arnab, Mostafa Dehghani, Georg Heigold, Chen Sun, Mario Lu{\v{c}}i{\'c},
  and Cordelia Schmid.
\newblock Vivit: A video vision transformer.
\newblock {\em arXiv preprint arXiv:2103.15691}, 2021.

\bibitem{beltagy2020longformer}
Iz Beltagy, Matthew~E Peters, and Arman Cohan.
\newblock Longformer: The long-document transformer.
\newblock {\em arXiv preprint arXiv:2004.05150}, 2020.

\bibitem{bertasius2021space}
Gedas Bertasius, Heng Wang, and Lorenzo Torresani.
\newblock Is space-time attention all you need for video understanding?
\newblock {\em arXiv preprint arXiv:2102.05095}, 2021.

\bibitem{carion2020end}
Nicolas Carion, Francisco Massa, Gabriel Synnaeve, Nicolas Usunier, Alexander
  Kirillov, and Sergey Zagoruyko.
\newblock End-to-end object detection with transformers.
\newblock In {\em European Conference on Computer Vision}, pages 213--229.
  Springer, 2020.

\bibitem{carreira2019short}
Joao Carreira, Eric Noland, Chloe Hillier, and Andrew Zisserman.
\newblock A short note on the kinetics-700 human action dataset.
\newblock {\em arXiv preprint arXiv:1907.06987}, 2019.

\bibitem{carreira2017quo}
Joao Carreira and Andrew Zisserman.
\newblock Quo vadis, action recognition? a new model and the kinetics dataset.
\newblock In {\em proceedings of the IEEE Conference on Computer Vision and
  Pattern Recognition}, pages 6299--6308, 2017.

\bibitem{kay2017kinetics}
J. {Carreira} and A. {Zisserman}.
\newblock Quo vadis, action recognition? a new model and the kinetics dataset.
\newblock In {\em 2017 IEEE Conference on Computer Vision and Pattern
  Recognition (CVPR)}, pages 4724--4733, July 2017.

\bibitem{chen2021empirical}
Xinlei Chen, Saining Xie, and Kaiming He.
\newblock An empirical study of training self-supervised visual transformers.
\newblock {\em arXiv preprint arXiv:2104.02057}, 2021.

\bibitem{chen2018multi}
Yunpeng Chen, Yannis Kalantidis, Jianshu Li, Shuicheng Yan, and Jiashi Feng.
\newblock Multi-fiber networks for video recognition.
\newblock In {\em Proceedings of the european conference on computer vision
  (ECCV)}, pages 352--367, 2018.

\bibitem{dai2020up}
Zhigang Dai, Bolun Cai, Yugeng Lin, and Junying Chen.
\newblock Up-detr: Unsupervised pre-training for object detection with
  transformers.
\newblock {\em arXiv preprint arXiv:2011.09094}, 2020.

\bibitem{devlin2018bert}
Jacob Devlin, Ming-Wei Chang, Kenton Lee, and Kristina Toutanova.
\newblock Bert: Pre-training of deep bidirectional transformers for language
  understanding.
\newblock {\em arXiv preprint arXiv:1810.04805}, 2018.

\bibitem{devlin2019bert}
Jacob Devlin, Ming-Wei Chang, Kenton Lee, and Kristina Toutanova.
\newblock Bert: Pre-training of deep bidirectional transformers for language
  understanding.
\newblock In {\em Proceedings of the 2019 Conference of the North American
  Chapter of the Association for Computational Linguistics: Human Language
  Technologies, Volume 1 (Long and Short Papers)}, pages 4171--4186, 2019.

\bibitem{dosovitskiy2020image}
Alexey Dosovitskiy, Lucas Beyer, Alexander Kolesnikov, Dirk Weissenborn,
  Xiaohua Zhai, Thomas Unterthiner, Mostafa Dehghani, Matthias Minderer, Georg
  Heigold, Sylvain Gelly, et~al.
\newblock An image is worth 16x16 words: Transformers for image recognition at
  scale.
\newblock {\em arXiv preprint arXiv:2010.11929}, 2020.

\bibitem{duke2021sstvos}
Brendan Duke, Abdalla Ahmed, Christian Wolf, Parham Aarabi, and Graham~W
  Taylor.
\newblock Sstvos: Sparse spatiotemporal transformers for video object
  segmentation.
\newblock {\em arXiv preprint arXiv:2101.08833}, 2021.

\bibitem{fan2021multiscale}
Haoqi Fan, Bo Xiong, Karttikeya Mangalam, Yanghao Li, Zhicheng Yan, Jitendra
  Malik, and Christoph Feichtenhofer.
\newblock Multiscale vision transformers.
\newblock {\em arXiv preprint arXiv:2104.11227}, 2021.

\bibitem{fan2019more}
Quanfu Fan, Chun-Fu Chen, Hilde Kuehne, Marco Pistoia, and David Cox.
\newblock More is less: Learning efficient video representations by big-little
  network and depthwise temporal aggregation.
\newblock {\em arXiv preprint arXiv:1912.00869}, 2019.

\bibitem{feichtenhofer2020x3d}
Christoph Feichtenhofer.
\newblock X3d: Expanding architectures for efficient video recognition.
\newblock In {\em Proceedings of the IEEE/CVF Conference on Computer Vision and
  Pattern Recognition}, pages 203--213, 2020.

\bibitem{feichtenhofer2018slowfast}
Christoph Feichtenhofer, Haoqi Fan, Jitendra Malik, and Kaiming He.
\newblock Slowfast networks for video recognition.
\newblock In {\em Proceedings of the IEEE International Conference on Computer
  Vision}, pages 6202--6211, 2019.

\bibitem{gabeur2020multi}
Valentin Gabeur, Chen Sun, Karteek Alahari, and Cordelia Schmid.
\newblock Multi-modal transformer for video retrieval.
\newblock In {\em European Conference on Computer Vision (ECCV)}, volume~5.
  Springer, 2020.

\bibitem{girdhar2019video}
Rohit Girdhar, Joao Carreira, Carl Doersch, and Andrew Zisserman.
\newblock Video action transformer network.
\newblock In {\em Proceedings of the IEEE/CVF Conference on Computer Vision and
  Pattern Recognition}, pages 244--253, 2019.

\bibitem{girdhar_CVPR2017_actionVLAD}
Rohit Girdhar, Deva Ramanan, Abhinav Gupta, Josef Sivic, and Bryan Russell.
\newblock {ActionVLAD: Learning Spatio-Temporal Aggregation for Action
  Classification}.
\newblock In {\em The IEEE Conference on Computer Vision and Pattern
  Recognition (CVPR)}, 2017.

\bibitem{goyal2017something}
Raghav Goyal, Samira~Ebrahimi Kahou, Vincent Michalski, Joanna Materzynska,
  Susanne Westphal, Heuna Kim, Valentin Haenel, Ingo Fruend, Peter Yianilos,
  Moritz Mueller-Freitag, et~al.
\newblock The" something something" video database for learning and evaluating
  visual common sense.
\newblock In {\em ICCV}, volume~1, page~3, 2017.

\bibitem{hara2018can}
Kensho Hara, Hirokatsu Kataoka, and Yutaka Satoh.
\newblock Can spatiotemporal 3d cnns retrace the history of 2d cnns and
  imagenet?
\newblock In {\em Proceedings of the IEEE conference on Computer Vision and
  Pattern Recognition}, pages 6546--6555, 2018.

\bibitem{hochreiter1997long}
Sepp Hochreiter and J{\"u}rgen Schmidhuber.
\newblock Long short-term memory.
\newblock {\em Neural computation}, 9(8):1735--1780, 1997.

\bibitem{ji20123d}
Shuiwang Ji, Wei Xu, Ming Yang, and Kai Yu.
\newblock 3d convolutional neural networks for human action recognition.
\newblock {\em IEEE transactions on pattern analysis and machine intelligence},
  35(1):221--231, 2012.

\bibitem{jiang_ICCV2019_STM}
Boyuan Jiang, MengMeng Wang, Weihao Gan, Wei Wu, and Junjie Yan.
\newblock {STM: SpatioTemporal and Motion Encoding for Action Recognition}.
\newblock In {\em The IEEE International Conference on Computer Vision (ICCV)},
  2019.

\bibitem{karpathy2014large}
Andrej Karpathy, George Toderici, Sanketh Shetty, Thomas Leung, Rahul
  Sukthankar, and Li Fei-Fei.
\newblock Large-scale video classification with convolutional neural networks.
\newblock In {\em Proceedings of the IEEE conference on Computer Vision and
  Pattern Recognition}, pages 1725--1732, 2014.

\bibitem{kuehne2011hmdb}
Hildegard Kuehne, Hueihan Jhuang, Est{\'\i}baliz Garrote, Tomaso Poggio, and
  Thomas Serre.
\newblock Hmdb: a large video database for human motion recognition.
\newblock In {\em 2011 International Conference on Computer Vision}, pages
  2556--2563. IEEE, 2011.

\bibitem{li2016action}
Qing Li, Zhaofan Qiu, Ting Yao, Tao Mei, Yong Rui, and Jiebo Luo.
\newblock Action recognition by learning deep multi-granular spatio-temporal
  video representation.
\newblock In {\em Proceedings of the 2016 ACM on International Conference on
  Multimedia Retrieval}, pages 159--166, 2016.

\bibitem{li2020nuta}
Xinyu Li, Chunhui Liu, Bing Shuai, Yi Zhu, Hao Chen, and Joseph Tighe.
\newblock Nuta: Non-uniform temporal aggregation for action recognition.
\newblock {\em arXiv preprint arXiv:2012.08041}, 2020.

\bibitem{li2020directional}
Xinyu Li, Bing Shuai, and Joseph Tighe.
\newblock Directional temporal modeling for action recognition.
\newblock In {\em European Conference on Computer Vision}, pages 275--291.
  Springer, 2020.

\bibitem{li_CVPR2020_TEA}
Yan Li, Bin Ji, Xintian Shi, Jianguo Zhang, Bin Kang, and Limin Wang.
\newblock {TEA: Temporal Excitation and Aggregation for Action Recognition}.
\newblock In {\em The IEEE Conference on Computer Vision and Pattern
  Recognition (CVPR)}, 2020.

\bibitem{li2020tea}
Yan Li, Bin Ji, Xintian Shi, Jianguo Zhang, Bin Kang, and Limin Wang.
\newblock Tea: Temporal excitation and aggregation for action recognition.
\newblock In {\em Proceedings of the IEEE/CVF Conference on Computer Vision and
  Pattern Recognition}, pages 909--918, 2020.

\bibitem{Li_CVPR2016_VLAD3}
Yingwei Li, Weixin Li, Vijay Mahadevan, and Nuno Vasconcelos.
\newblock {VLAD3: Encoding Dynamics of Deep Features for Action Recognition}.
\newblock In {\em The IEEE Conference on Computer Vision and Pattern
  Recognition (CVPR)}, 2016.

\bibitem{li2020bridging}
Zekang Li, Zongjia Li, Jinchao Zhang, Yang Feng, Cheng Niu, and Jie Zhou.
\newblock Bridging text and video: A universal multimodal transformer for
  video-audio scene-aware dialog.
\newblock {\em arXiv preprint arXiv:2002.00163}, 2020.

\bibitem{lin2019tsm}
Ji Lin, Chuang Gan, and Song Han.
\newblock Tsm: Temporal shift module for efficient video understanding.
\newblock In {\em Proceedings of the IEEE/CVF International Conference on
  Computer Vision}, pages 7083--7093, 2019.

\bibitem{liu2021swin}
Ze Liu, Yutong Lin, Yue Cao, Han Hu, Yixuan Wei, Zheng Zhang, Stephen Lin, and
  Baining Guo.
\newblock Swin transformer: Hierarchical vision transformer using shifted
  windows.
\newblock {\em arXiv preprint arXiv:2103.14030}, 2021.

\bibitem{liu_AAAI2020_TEINet}
Zhaoyang Liu, Donghao Luo, Yabiao Wang, Limin Wang, Ying Tai, Chengjie Wang,
  Jilin Li, Feiyue Huang, and Tong Lu.
\newblock {TEINet: Towards an Efficient Architecture for Video Recognition}.
\newblock In {\em The Conference on Artificial Intelligence (AAAI)}, 2020.

\bibitem{liu2021video}
Ze Liu, Jia Ning, Yue Cao, Yixuan Wei, Zheng Zhang, Stephen Lin, and Han Hu.
\newblock Video swin transformer.
\newblock {\em arXiv preprint arXiv:2106.13230}, 2021.

\bibitem{neimark2021video}
Daniel Neimark, Omri Bar, Maya Zohar, and Dotan Asselmann.
\newblock Video transformer network.
\newblock {\em arXiv preprint arXiv:2102.00719}, 2021.

\bibitem{patrick2021keeping}
Mandela Patrick, Dylan Campbell, Yuki~M Asano, Ishan Misra~Florian Metze,
  Christoph Feichtenhofer, Andrea Vedaldi, Jo Henriques, et~al.
\newblock Keeping your eye on the ball: Trajectory attention in video
  transformers.
\newblock {\em arXiv preprint arXiv:2106.05392}, 2021.

\bibitem{piergiovanni2019tiny}
AJ Piergiovanni, Anelia Angelova, and Michael~S Ryoo.
\newblock Tiny video networks.
\newblock {\em arXiv preprint arXiv:1910.06961}, 2019.

\bibitem{shao2020temporal}
Hao Shao, Shengju Qian, and Yu Liu.
\newblock Temporal interlacing network.
\newblock In {\em Proceedings of the AAAI Conference on Artificial
  Intelligence}, volume~34, pages 11966--11973, 2020.

\bibitem{sigurdsson2016hollywood}
Gunnar~A. Sigurdsson, G{\"u}l Varol, Xiaolong Wang, Ivan Laptev, Ali Farhadi,
  and Abhinav Gupta.
\newblock Hollywood in homes: Crowdsourcing data collection for activity
  understanding.
\newblock {\em ArXiv e-prints}, 2016.

\bibitem{soomro2012dataset}
Khurram Soomro, Amir~Roshan Zamir, and M Shah.
\newblock A dataset of 101 human action classes from videos in the wild.
\newblock {\em Center for Research in Computer Vision}, 2012.

\bibitem{touvron2020training}
Hugo Touvron, Matthieu Cord, Matthijs Douze, Francisco Massa, Alexandre
  Sablayrolles, and Herv{\'e} J{\'e}gou.
\newblock Training data-efficient image transformers \& distillation through
  attention.
\newblock {\em arXiv preprint arXiv:2012.12877}, 2020.

\bibitem{tran2015learning}
Du Tran, Lubomir Bourdev, Rob Fergus, Lorenzo Torresani, and Manohar Paluri.
\newblock Learning spatiotemporal features with 3d convolutional networks.
\newblock In {\em Proceedings of the IEEE international conference on computer
  vision}, pages 4489--4497, 2015.

\bibitem{tran2019video}
Du Tran, Heng Wang, Lorenzo Torresani, and Matt Feiszli.
\newblock Video classification with channel-separated convolutional networks.
\newblock In {\em Proceedings of the IEEE/CVF International Conference on
  Computer Vision}, pages 5552--5561, 2019.

\bibitem{tran2018closer}
Du Tran, Heng Wang, Lorenzo Torresani, Jamie Ray, Yann LeCun, and Manohar
  Paluri.
\newblock A closer look at spatiotemporal convolutions for action recognition.
\newblock In {\em Proceedings of the IEEE conference on Computer Vision and
  Pattern Recognition}, pages 6450--6459, 2018.

\bibitem{ullah2017action}
Amin Ullah, Jamil Ahmad, Khan Muhammad, Muhammad Sajjad, and Sung~Wook Baik.
\newblock Action recognition in video sequences using deep bi-directional lstm
  with cnn features.
\newblock {\em IEEE access}, 6:1155--1166, 2017.

\bibitem{vaswani2017attention}
Ashish Vaswani, Noam Shazeer, Niki Parmar, Jakob Uszkoreit, Llion Jones,
  Aidan~N Gomez, Lukasz Kaiser, and Illia Polosukhin.
\newblock Attention is all you need.
\newblock {\em arXiv preprint arXiv:1706.03762}, 2017.

\bibitem{wang2020video}
Heng Wang, Du Tran, Lorenzo Torresani, and Matt Feiszli.
\newblock Video modeling with correlation networks.
\newblock In {\em Proceedings of the IEEE/CVF Conference on Computer Vision and
  Pattern Recognition}, pages 352--361, 2020.

\bibitem{wang_ECCV2016_TSN}
Limin Wang, Yuanjun Xiong, Zhe Wang, Yu Qiao, Dahua Lin, Xiaoou Tang, and Luc
  Van~Gool.
\newblock {Temporal Segment Networks: Towards Good Practices for Deep Action
  Recognition}.
\newblock In {\em The European Conference on Computer Vision (ECCV)}, 2016.

\bibitem{wang2018non}
Xiaolong Wang, Ross Girshick, Abhinav Gupta, and Kaiming He.
\newblock Non-local neural networks.
\newblock In {\em Proceedings of the IEEE conference on computer vision and
  pattern recognition}, pages 7794--7803, 2018.

\bibitem{wu2019long}
Chao-Yuan Wu, Christoph Feichtenhofer, Haoqi Fan, Kaiming He, Philipp
  Krahenbuhl, and Ross Girshick.
\newblock Long-term feature banks for detailed video understanding.
\newblock In {\em Proceedings of the IEEE/CVF Conference on Computer Vision and
  Pattern Recognition}, pages 284--293, 2019.

\bibitem{wu2021cvt}
Haiping Wu, Bin Xiao, Noel Codella, Mengchen Liu, Xiyang Dai, Lu Yuan, and Lei
  Zhang.
\newblock Cvt: Introducing convolutions to vision transformers.
\newblock {\em arXiv preprint arXiv:2103.15808}, 2021.

\bibitem{xie2017aggregated}
Saining Xie, Ross Girshick, Piotr Doll{\'a}r, Zhuowen Tu, and Kaiming He.
\newblock Aggregated residual transformations for deep neural networks.
\newblock In {\em Proceedings of the IEEE conference on computer vision and
  pattern recognition}, pages 1492--1500, 2017.

\bibitem{yang_CVPR2020_TPN}
Ceyuan Yang, Yinghao Xu, Jianping Shi, Bo Dai, and Bolei Zhou.
\newblock {Temporal Pyramid Network for Action Recognition}.
\newblock In {\em The IEEE Conference on Computer Vision and Pattern
  Recognition (CVPR)}, 2020.

\bibitem{yang2020temporal}
Ceyuan Yang, Yinghao Xu, Jianping Shi, Bo Dai, and Bolei Zhou.
\newblock Temporal pyramid network for action recognition.
\newblock In {\em Proceedings of the IEEE/CVF Conference on Computer Vision and
  Pattern Recognition}, pages 591--600, 2020.

\bibitem{yang2020transpose}
Sen Yang, Zhibin Quan, Mu Nie, and Wankou Yang.
\newblock Transpose: Towards explainable human pose estimation by transformer.
\newblock {\em arXiv preprint arXiv:2012.14214}, 2020.

\bibitem{yuan2021tokens}
Li Yuan, Yunpeng Chen, Tao Wang, Weihao Yu, Yujun Shi, Francis~EH Tay, Jiashi
  Feng, and Shuicheng Yan.
\newblock Tokens-to-token vit: Training vision transformers from scratch on
  imagenet.
\newblock {\em arXiv preprint arXiv:2101.11986}, 2021.

\bibitem{yue2015beyond}
Joe Yue-Hei~Ng, Matthew Hausknecht, Sudheendra Vijayanarasimhan, Oriol Vinyals,
  Rajat Monga, and George Toderici.
\newblock Beyond short snippets: Deep networks for video classification.
\newblock In {\em Proceedings of the IEEE conference on computer vision and
  pattern recognition}, pages 4694--4702, 2015.

\bibitem{zhou_ECCV2018_TRN}
Bolei Zhou, Alex Andonian, Aude Oliva, and Antonio Torralba.
\newblock {Temporal Relational Reasoning in Videos}.
\newblock In {\em The European Conference on Computer Vision (ECCV)}, 2018.

\bibitem{zhou2018end}
Luowei Zhou, Yingbo Zhou, Jason~J Corso, Richard Socher, and Caiming Xiong.
\newblock End-to-end dense video captioning with masked transformer.
\newblock In {\em Proceedings of the IEEE Conference on Computer Vision and
  Pattern Recognition}, pages 8739--8748, 2018.

\end{thebibliography}
}




\end{document}